\documentclass[oneside,12pt]{IIScthesisPSnPDF}
\usepackage{amssymb}
\usepackage{amsmath}
\usepackage{latexsym}
\usepackage{multicol}
\usepackage{booktabs}
\usepackage{multirow}
\usepackage{fancyhdr}
\usepackage[numbers]{natbib}
\usepackage{cleveref}
\usepackage{wrapfig}
\usepackage{algorithm}
\usepackage{algorithmic}
\usepackage{enumerate}
\usepackage{cancel}
\usepackage{mathrsfs}
\usepackage{color}
\usepackage[compact]{titlesec}
\usepackage{mdwlist}
\usepackage{tikz}
\usepackage{varwidth}
\usepackage[vertfit]{breakurl}
\usepackage{datetime}
\usepackage{pdfpages}
\usetikzlibrary{shapes,arrows, trees}

\usepackage{subcaption}

\newtheorem{theorem}{Theorem}[chapter]
\newtheorem{lemma}{Lemma}[chapter]

\newcommand{\remove}[1]{}

\newcommand{\notes}[1]{}

\newcommand{\first}[1]{$1^{\mathrm{st}}$}
\newcommand{\second}[1]{$2^{\mathrm{nd}}$}

\newcommand{\squishlisttwo}{
\begin{list}{$\blacktriangleright$}
{ \setlength{\itemsep}{0.5pt}
\setlength{\parsep}{0pt}
\setlength{\topsep}{0pt}
\setlength{\partopsep}{0.5pt}
\setlength{\leftmargin}{1em}
\setlength{\labelwidth}{1em}
\setlength{\labelsep}{0.5em} } }

\newcommand{\squishend}{
\end{list} }
\allowdisplaybreaks[1]

\newdateformat{monthyeardate}{ \monthname[\THEMONTH], \THEYEAR}

\newcommand{\blankpage}{
\newpage
\thispagestyle{empty}
\mbox{}
\newpage
}

\crefname{observation}{observation}{observations}
\crefname{algorithm}{algorithm}{algorithms}
\crefname{align}{equation}{equations}
\crefname{eqnarray}{equation}{equations}

\begin{document}

\title{The IISc Thesis Template} 

\submitdate{\monthyeardate\today} 
\mtech
\dept{Computer Science and Automation}
\faculty{Faculty of Engineering}
\author{Samineni Soumya Rani}


\maketitle

\begin{center}
\LARGE{\underline{\textbf{Declaration of Originality}}}
\end{center}
\noindent I, \textbf{Samineni Soumya Rani}, with SR No. \textbf{04-04-00-10-42-19-1-17066} hereby declare that
the material presented in the thesis titled

\begin{center}
\textbf{Policy Search using Dynamic Mirror Descent MPC for Model Free RL}
\end{center}

\noindent represents original work carried out by me in the \textbf{Department of Computer Science and Automation} at \textbf{Indian Institute of Science} during the years \textbf{2019-2021}.

\noindent With my signature, I certify that:
\begin{itemize}
	\item I have not manipulated any of the data or results.
	\item I have not committed any plagiarism of intellectual
	property.
	I have clearly indicated and referenced the contributions of
	others.
	\item I have explicitly acknowledged all collaborative research
	and discussions.
	\item I have understood that any false claim will result in severe
	disciplinary action.
	\item I have understood that the work may be screened for any form
	of academic misconduct.
\end{itemize}

\vspace{20mm}

\noindent {\footnotesize{Date:	\hfill	Student Signature}} \qquad

\vspace{20mm}

\noindent In my capacity as supervisor of the above-mentioned work, I certify
that the above statements are true to the best of my knowledge, and 
I have carried out due diligence to ensure the originality of the
report.

\vspace{20mm}

\noindent  {\footnotesize{Advisor Name:\hfill Advisor Signature \qquad}\newline Dr.Shishir Nadubettu Yadukumar Kolathaya \newline  Dr.Shalabh Bhatnagar }

\blankpage

\vspace*{\fill}
\begin{center}
\large\bf \textcopyright \ Samineni Soumya Rani\\
\large\bf \monthyeardate\today\\
\large\bf All rights reserved
\end{center}
\vspace*{\fill}
\thispagestyle{empty}

\blankpage

\vspace*{\fill}
\begin{center}
DEDICATED TO \\[2em]
\Large\it My beloved father\\[2em]
\Large\it who has always been my inspiration.
\end{center}
\vspace*{\fill}
\thispagestyle{empty}




\setcounter{secnumdepth}{3}
\setcounter{tocdepth}{3}

\frontmatter 
\pagenumbering{roman}

\prefacesection{Acknowledgements}
I would like to express my gratitude to Prof.Shishir Nadubettu Yadukumar Kolathaya and Prof.Shalabh Bhatnagar for the opportunity to work under their guidance and for helping me throughout my M.Tech, early from second semester to writing my first paper based on this Project. You both have always been available whenever I needed help. While I was juggling with many research directions you helped me to stay focused. The freedom you offered me in exploring areas of Reinforcement Learning and Robotics had a great impact on my graduate research experience. This work wouldn’t have been possible without the constant support and help from both of you.\\
\\
I would also like to thank Prof.Aditya Gopalan for sharing his valuable feedback in evaluating my thesis and for his course on Online Learning.\\ 
\\
A special thank you goes out to Utkarsh Aashu Mishra for helping me especially for being a good company.\\ 
\\
I would like to thank my mother for always motivating and encouraging me to acheive what I beleive in and I would also like to thank my brother and my sister in law for their support during my stay in Bengluru during Covid.\\ 
\\
Further, I thank all the members of Stoch Lab and Stochastic Systems Lab for making the experience a pleasant one. The exposure I got through both the labs was invaluable.\\ 
\\ I would like to thank all the wonderful friends I made here who made my Journey at IISc the most memorable one and I cherish all the moments we had together.   

\prefacesection{Abstract}
Recent works in Reinforcement Learning (RL) combine model-free (Mf)-RL algorithms with model-based (Mb)-RL approaches to get the best from both: asymptotic performance of Mf-RL and high sample-efficiency of Mb-RL. Inspired by these works, we propose a hierarchical framework that integrates online learning for the Mb-trajectory optimization with off-policy methods for the Mf-RL. In particular, two loops are proposed, where the Dynamic Mirror Descent based Model Predictive Control (DMD-MPC) is used as the inner loop to obtain an optimal sequence of actions. These actions are in turn used to significantly accelerate the outer loop Mf-RL. We show that our formulation is generic for a broad class of MPC based policies and objectives, and includes some of the well-known Mb-Mf approaches. Based on the frame work we define two algorithms to increase sample efficiency of Off Policy RL and to guide end to end RL algorithms for online adaption respectively. Thus we finally introduce two novel algorithms: \emph{\textbf D}ynamic-Mirror D\emph{\textbf e}scent \emph{\textbf {Mo}}del Predictive \emph{\textbf {RL}} (\textbf{DeMoRL}), which uses the method of elite fractions for the inner loop and Soft Actor-Critic (SAC) as the off-policy RL for the outer loop and \emph{\textbf D}ynamic-Mirror D\emph{\textbf e}scent \emph{\textbf {Mo}}del Predictive \emph{\textbf {Layer}} (\textbf{DeMo Layer}), a special case of hierarchical framework which guides linear policies trained using Augmented Random Search(ARS). Our experiments show faster convergence of the proposed DeMo RL, and better or equal performance compared to other Mf-Mb approaches on benchmark MuJoCo control tasks. The DeMo Layer was tested on classical Cartpole and custom built Quadruped trained using Linear Policy Approach. The results shows that DeMo Layer significantly increases performance of the Linear Policy in both the settings.

\prefacesection{Acronyms}
\begin{table}[h]
\caption{Key Acronyms used in the Report}
\label{table}
\centering
\begin{tabular}{l  l}
\hline
\hline
\textbf{Acronyms} & \textbf{Expansion} \\
\hline
DMD MPC & Dynamic Mirror Descent Model Predictive Control  \\
\hline
ARS & Augmented Random Search \\
\hline
SAC & Soft Actor Critic \\
\hline
DeMo RL & Dynamic Mirror Descent Model Predictive RL\\
\hline
MoPAC & Model Predictive Actor Critic \\
\hline
TOPDM & Trajectory Optimisation for Precise Dexterous
Manipulation \\
\hline
\hline
\end{tabular}
\end{table}

\prefacesection{Publications based on this Thesis}
 Accelerating Actor-Critic with Dynamic Mirror
Descent based Model Predictive Control (Under Review) submitted to CoRL 2021

\tableofcontents
\listoffigures
\listoftables

\mainmatter 
\setcounter{page}{1}
\chapter{Introduction}
\label{chap:introduction}



Deep Reinforcement Learning (DRL) algorithms are shown to be highly successful in challenging control tasks like dexterous manipulation \cite{Rajeswaran-RSS-18}, agile locomotion \cite{peng2020learning,lee2020learning} and recent works demonstrate applications to safety critical systems as well \cite{dalal2018safe,zhang2020cautious,robey2021learning}. 
But the PROBLEMS with Deep RL techniques are: 
\begin{enumerate}
     \item Policies trained in simulation often fail to transfer to real hardware. 
     \item Any modification/update in the model renders the previously trained policy invalid. In other words, the policies are sensitive to changes in the model.
     \item Training in hardware is very expensive and time consuming.Further, any undesirable behavior may lead to damages in the hardware.
\end{enumerate}
Additional problems involve, the lack of model information in training Model Free RL algorithms, which results in prolonged training. Hence, DRL algorithms need to be made Data efficient, Generalisable and further have to adapt to unseen environments.Adapting as efficiently as possible requires perceiving the environment and an investment in learning the Model is highly sample efficient, results in Model Based RL techniques. But Models fail to capture the Global Dynamics and Model Based RL Techniques fail to achieve asymptotic performance of Model Free RL Techniques. Several works combined both the Model Based and Model Free Approaches and we propose a generalised framework combining the Model Based MPC with Model Free Off Policy RL algorithms.


\section{Motivation}
Model-Free Reinforcement Learning (Mf-RL) algorithms are widely applied to solve challenging control tasks as they eliminate the need to model the complex dynamics of the system. However, these techniques are significantly data hungry and require millions of transitions. Furthermore, these characteristics highly limit successful training on hardware as undergoing such high number of transitions in hardware environments is infeasible. Thus, in order to overcome this hurdle, various works have settled for a two loop model-based approach, typically referred to as Model-based Reinforcement Learning (Mb-RL) algorithms. Such strategies take the benefit of the explored dynamics of the system by learning the dynamics model, and then determining an optimal policy in this model. Hence this ``inner-loop" optimization allows for a better choice of actions
before interacting  with the original environment.
\\

The inclusion of model-learning in RL has significantly improved sampling efficiency \cite{Levine13GPS,nagabandi2018neural}, and there are numerous works in this direction.
DRL algorithms, while exploring, collect significant amount of state transitions, which can be used to generate an approximate dynamics model of the system. 
In the context of robotics, this model has proven to be very beneficial in developing robust control strategies based on predictive simulations \cite{NEURIPS2019_MBPO}. They have successfully handled minor disturbances and demonstrated sim2real feasibility. 
Moreover, the process of planning with the learnt model is mainly motivated  by the Model Predictive Control (MPC), which is a well known strategy used in classical real-time control. Given the model and the cost formulation, a typical MPC structure can be formulated in the form of a finite horizon trajectory optimization problem. Thus our work is motivated to propose a generalised framework combining Model Free and Model Based methods.

\section{Related Work}
 The work with such a view of Mb-Mf approach and exploiting the approximated dynamics with random shooting, \cite{nagabandi2018neural} demonstrated its efficacy in leveraging the overall learning performance. Further, the work also showed how model-based (Mb) additions to typical model-free (Mf) algorithms can accelerate significantly the latter ones. Additionally, in this context of Mb-Mf RL algorithms, \cite{lowrey2018plan} also introduced the use of value functions with an MPC formulation and \cite{hafner2018planet} shows a similar formulation with high-dimensional image observations. Recent works presented in \cite{NEURIPS2020_MOPO} showed adaptation to Dynamic changes using MPC with world models and \cite{morgan2021model} proposes an actor critic framework using model predictive rollouts and demonstrated applicability on real hardware. The TOPDM \cite{Charlesworth2020SolvingCD}, a close approach to DeMo RL demonstrates spinning a pen between the fingers, the most challenging examples in dexterous hand manipulation.
\\

Further, prior works \cite{Levine13GPS}, \cite{Schulman2015TRPO}, \cite{Tomar2020MDPO} have explored guiding RL Policies using Mirror Descent Approaches with KL Constraint on the policy update. As far as our knowledge, we are the first to generalise the Mb Mf Framework in the literature with the view of Dynamic Mirror Descent MPC to RL polices

\section{Contribution}
With a view toward strengthening existing Mb-Mf approaches for learning, we propose a generic framework that integrates
a model-based optimization scheme with model-free off-policy learning.
Motivated by the success of online learning algorithms \cite{wagener2019online} in RC buggy models, we combine them with off-policy Mf learning, thereby leading to a two-loop Mb-Mf approach.
%
In particular, we implement dynamic mirror descent (DMD) algorithms on a model-estimate of the system, and then the outer loop Mf-RL is used on the real system. 
The main advantage with this setting is that the inner loop is computationally light; the number of iterations can be large without effecting the overall performance. 
%
%
Since this is a hierarchical approach, the inner loop policy helps improve the outer loop policy, by effectively utilizing the control choices made on the approximate dynamics. This approach, in fact, provides a more generic framework for some of the Mb-Mf approaches (e.g., \cite{morgan2021model}, \cite{nagabandi2020deep}).
\\

In addition to the proposed framework, we introduce two new algorithms DeMo RL and DeMo Layer. The Dynamic Mirror-Descent Model Predictive RL (DeMoRL), 
 uses Soft actor-critic (SAC) \cite{haarnoja2018soft} in the outer loop as off-policy RL, and Cross-Entropy Method (CEM) in the inner loop as DMD-MPC \cite{wagener2019online}.
In particular, we use the exponential family of control distributions with CEM on the objective. 
In each iteration, the optimal control sequence obtained is then applied on the model-estimate to collect additional data.
This is appended to the buffer, which is then used by the outer-loop for learning the optimal policy. We show that the DMD-MPC accelerates the learning of the outer-loop by simply enriching the data with better choices of state-control transitions. We finally demonstrate this method on custom robotic environments and MuJoCo benchmark control tasks.
Simulation results show that the proposed methodology is better than or at least as good as MoPAC \cite{morgan2021model} and MBPO \cite{NEURIPS2019_MBPO} in terms of sample-efficiency.
Furthermore, as our formulation is closer to that of \cite{morgan2021model}, it is worth mentioning that even though we do not show results in hardware, the proposed algorithms can be used to train in hardware more effectively, which will be a part of future work.
\\

The DeMo Layer, a special instance of hierarchical framework guides linear policies trained using Augmented Random Search(ARS). The experiments are conducted Cartpole swing up and quadrupedal walking. Our experimental results show that proposed DeMo Layer could improve the policy and could be used end to end with any RL algorithm during deployment.


\section{Outline of the Report}
The report is structured as follows: 
\begin{itemize}
    \item Chapter \ref{sec:background}. \textbf{Problem Formulation}\\
    In this chapter, we provide the preliminaries for Reinforcement Learning and Online Learning as followed in the report. We further describe the RL algorithms in specific the Augmented Random Search, Soft Actor Critic and Online Learning approach to MPC.
    \item Chapter \ref{sec:methodology}. \textbf{Methodology: Novel Framework \& Algorithms} \\ 
    We will describe the hierarchical framework for the proposed strategy, followed by the description of the DMD-MPC. With the proposed generalised framework, we formulate the two novel algorithms associated with the strategy DeMo RL and DeMO Layer in this chapter.
    \item Chapter \ref{sec:result}. \textbf{Experimental Results} \\
    In this chapter we run experiments of DeMo RL on benchmark Mujoco Control Tasks and we compare the results with existing and state of the art algorithms MOPAC and MBPO. The experiments of DeMO Layer was conducted on swinging up Cartpole and custom built quadruped Stoch2.  Further, we discuss our experimental results and show significance of proposed algorithms.
    \item Chapter \ref{sec:conclusion}. \textbf{Conclusion \& Future Work}\\ Finally, we end the report by summarizing the work done and proposing some interesting future directions.

\end{itemize}

\chapter{Preliminaries}
\label{sec:background}
\section{Optimal Control- MPC}
The Model Predictive Control is a widely applied control strategy and gives practical and robust controllers. It considers a stochastic dynamics model $\hat{f}$ an approximation to real system $f$ and solves an H step optimisation problem at every time step and applies first control to the real dynamical system $f$ to go to the next state $x_{t+1}$. 
            A popular MPC objective is the expected $H$ -step future costs
            \begin{equation}
            {J}\left( x_{t}\right)=\mathbb{E}\left[C\left({\boldsymbol{x}}_{t}, {\boldsymbol{u}}_{t}\right) \right],
            \end{equation}
            \begin{align}
    C\left(\mathbf{x_t}, \mathbf{u_t}\right) = \sum^{H-1}_{h = 0} \gamma^h c(x_{t,h},u_{t,h}) + \gamma^H c_{H}(x_{t,H}) 
\end{align}
where, $c(x_{t,h}, u_{t,h})$ is the cost incurred (for the control problem) and $c_{H}(x_{t,H})$ is the terminal cost.
            \\
            Since optimal control is obtained from ${J}\left( x_{t}\right),$ which is based on $x_{t}$, thus MPC is effectively state-feedback as desired for a stochastic system and is an effective tool for control tasks involving dynamic environments or non stationary setup.\\
            \\
            Though MPC sounds intuitively promising, the optimization is approximated in practice and the control command $u_{t}$ needs to be computed in real time at high frequency. Hence, a common practice is to heuristically bootstrap the previous approximate solution as the initialization to the current problem. 
 
\section{Reinforcement Learning Framework}

We consider an infinite horizon Markov Decision Process (MDP) given by $\{\mathcal{X}, \mathcal{U}, r, P, \gamma, \rho_0 \}$ where $\mathcal{X}~\subset~\mathbb{R}^n$ refers to set of states of the robot and $\mathcal{U}~\subset~\mathbb{R}^m$ refers to the set of control or  actions. $r: \mathcal{X} \times \mathcal{U} \rightarrow \mathbb{R}$ is the reward function, $P : \mathcal{X} \times \mathcal{U} \times \mathcal{X} \rightarrow [0,1]$ refers to the function that gives transition probabilities between two states for a given action, and $\gamma \in (0,1)$ is the discount factor of the MDP. The distribution over initial states is given by $\rho_0: \mathcal{X} \rightarrow [0,1]$ and the policy is represented by $\pi_\theta: \mathcal{X} \to \mathcal{U}$ parameterized by $\theta \in \Theta$, a potentially feasible high-dimensional space. 
If a stochastic policy is used, then $\pi_\theta:\mathcal{X}\times \mathcal{U}\to [0,1]$. 
For ease of notations, we will use a deterministic policy to formulate the problem. Wherever a stochastic policy is used, we will show the extensions explicitly. 
In this formulation, the optimal policy is the policy that maximizes the expected return ($R$):
\begin{equation*}
R = \mathbb{E}[r_t+ \gamma r_{t+1} + \gamma^2 r_{t+2} + \dots]
\end{equation*}
where the subscript for $r_t$ denotes the step index. Note that the system model dynamics can be expressed in the form of an equation:
\begin{equation*}
    x_{t+1}  \sim f(x_t,u_t),
\end{equation*}

\begin{equation}
    \theta^* := \arg \max_{\theta} \mathbb{E}_{ \rho_0, \pi_\theta} \left[ \sum^{\infty}_{t = 0} \gamma^t r(x_t, u_t) \right], \quad x_0 \sim \rho_0, \quad x_{t+1} \sim f(x_t,\pi_\theta(x_t)).
\end{equation}
The offpolicy techniques like TD3, SAC have shown better sample complexity compared to TRPO, PPO. A simple random search based a Model Free Technique, Augmented Random Search \cite{mania2018simple}, proposed a Linear deterministic  policy highly competitive to other Model Free RL Techniques like TRPO, PPO and SAC. In the subsequent sections we describe the ARS algorithm in detail along with the improvement in its implementation and we also describe SAC. 

\section{Online Learning Framework}
            Another sequential decision making technique, Online learning is a framework for analyzing online decision making, essentially with three components: the decision set, the learner’s strategy for updating decisions, and the environment’s strategy for updating per-round losses.\\
            \\
            At round $t,$ the learner makes a decision $\tilde{\boldsymbol{\theta}}_{t}$,along with a side information $u_{t-1}$, then environment chooses a loss function $\ell_{t}$ and the learner suffers a cost $\ell_{t}\left(\tilde{\boldsymbol{\theta}}_{t}\right)$. along with side information like the gradient of loss to aid in choosing the next decision. \\
            \\
            Here, the learner's goal is to minimize the accumulated costs
            $\sum_{t=1}^{T} \ell_{t}\left(\tilde{\boldsymbol{\theta}}_{t}\right),$ i.e., by minimizing the regret.
            We describe, in detail the Online Learning Approach to Model Predictive Control  \cite{wagener2019online} in subsequent sections.

\section{Description of Algorithms}
    We describe the RL and Online Learning algorthms that are used in this work - Augmented Random Search(ARS), Soft Actor Critic(SAC) and Online Learning Approach to MPC.
        \subsection{Augmented Random Search}
            Random Search is a Derivative Free Optimisation where the gradient is estimated through finite difference Method \cite{nesterov2017random}.Objective is to maximize Expected return of a policy $\pi$ parameterised by $\theta$ under noise $\xi$
            \[\max _{\theta} \mathbb{E}_{\xi}\left[r\left(\pi_{\theta}, \xi\right)\right]\]
            The gradient is found from the gradient estimate obtained from gradient of smoothened version of above objective with Gaussian noise unlike from policy gradient theorem. Gradient of smoothened objective is \[ \frac{r\left(\pi_{\theta+\nu \delta}, \xi_{1}\right)-r\left(\pi_{\theta}, \xi_{2}\right)}{\nu}\delta \] where $\delta$ is zero mean Gaussian. If $\nu$ is sufficiently small, the Gradient estimate would be close to the gradient of original objective. Further bias could be reduced with a two point estimate, \[\frac{r\left(\pi_{\theta+\nu \delta}, \xi_{1}\right)-r\left(\pi_{\theta-\nu \delta}, \xi_{2}\right)}{\nu}\delta.\]
            A Basic Random Search would involve the update of policy parameters according to
            \begin{equation}
                \theta_{j+1}=\theta_{j}+\frac{\alpha}{N} \sum_{k=1}^{N}\left[r\left(\pi_{j, k,+}\right)-r\left(\pi_{j, k,-}\right)\right] \delta_{k}
            \end{equation}
            Augmented Random Search, defines an update rule, 
            \begin{equation}
                  \theta_{j+1}=\theta_{j}+\frac{\alpha}{b \sigma_{R}} \sum_{k=1}^{b}\left[r\left(\pi_{j,(k),+}\right)-r\left(\pi_{j,(k),-}\right)\right] \delta_{(k)}
            \end{equation}
            Policy is linear state feedback law, \[pi_{j}(x)=\left(\theta_{j}\right)\left(x\right)\]
            where x is the state and It proposes three Augmentations to Basic Random Search.

            i) Using top best b performing directions,
            They order the perturbation directions $+\delta_{(k)}$, in decreasing order according to $\max r\left(\pi_{j,(k),+}\right)$ and $r\left(\pi_{j,(k),-}\right)$ and uses only the top b directions.\\
            ii)Scaling by the standard deviation, helps in an adjusting the step size. \\
            iii) Normalization of the states
            \[\pi_{j}(x)=\left(\theta_{j}\right) \operatorname{diag}\left(\Sigma_{j}\right)^{-1 / 2}\left(x-\mu_{j}\right)
\]
            
            \subsubsection*{Accelerating ARS}
                Most optimisers use Adam to accelerate Stochastic Gradient Descent \cite{jain2017accelerating} in practical implementations. Hence with ARS we estimate the gradient, an acceleration technique is not used. So, we define an acceleration based Gradient Estimate to ARS for faster convergence. Future Work would involve validating this approach.
                The Modified ARS Algorithm with $\alpha$ and $\beta$ are the small and large step sizes respectively.
                \\
                \begin{figure}[ht]
                    \centering
                \begin{minipage}{.6\linewidth}
                \begin{algorithm}[H]
                    \caption{Accelerated ARS}
                    \label{arsalgorithm}
                    \begin{algorithmic}[1]
                    \STATE $\text Runaverage_{j} = \sum_{i< j} \left(1-\beta \right) ^i \theta_{\left(i-\tau \right)}$
                    \STATE $\theta_{j+1}=\theta_{j}+\frac{\alpha}{b \sigma_{R}} \sum_{k=1}^{b}\left[r\left(\pi_{j,(k),+}\right)-r\left(\pi_{j,(k),-}\right)\right] \delta_{(k)}$
                    \STATE $\theta_{acc_{j+1}} = \gamma \theta_{j+1}+ \left(1-\gamma \right) Runaverage_{j}$
                    \end{algorithmic}
                \end{algorithm}
                \end{minipage}
                \end{figure}
\subsection{Soft Actor Critic}
Soft Actor-Critic (SAC) \cite{haarnoja2018soft} is an offpolicy model-free RL algorithm based on principle of entropy maximization, with entropy of policy in addition to reward. 
It uses soft policy iteration for policy evaluation and improvement. It uses two Q Value functions to mitigate positive bias of value based methods and a minimum of the Q-functions is used for the value gradient and policy gradient. Further, two Q Functions Speeds up training process.
It also uses a target network with weights updated by exponentially moving average, with a smoothing constant $tau$, to increase stability. \\
The SAC policy $\pi_\theta$ is updated using the loss function
\begin{equation*}
    J_\theta = \mathbb{E}_{(x,u,r,x')\sim D}[D_{KL}(\pi||\exp{(Q_\xi - V_\zeta)})]
\end{equation*} where $D$, $V_\zeta$ and $Q_\xi$ represent the replay buffer, value function and Q-function associated with $\pi_\theta$. 
The exploration by SAC helps in learning the underlying dynamics. 
In each gradient step we  update SAC parameters using data 
\begin{equation*}
    \zeta \leftarrow \zeta-\lambda_{\psi} \nabla_{\zeta} J_{V_{\zeta}}
\end{equation*}
\begin{equation*}
    \xi \leftarrow \xi-\lambda_{\xi} \nabla_{\xi} J_{Q_{\xi}}
\end{equation*}
\begin{equation*}
    \theta_{-} \leftarrow \theta-\lambda_{\theta} \nabla_{\theta} J_{\pi_{\theta}}
\end{equation*}
\begin{equation*}
    \bar{\zeta} \leftarrow \tau \zeta+(1-\tau) \zeta
\end{equation*}
\begin{equation*}
    \bar{\xi} \leftarrow \tau \xi+(1-\tau) \xi
\end{equation*}
$\bar{\xi}$ and $\bar{\zeta}$ represent target networks.

\subsection{Online Learning for MPC}

The Online Learning (OL) makes a decision at time $t$ to optimise for the regret over time while MPC also optimizes for a finite $H$-step horizon cost at every time instant, thus having a close similarity to OL \cite{wagener2019online}. \\
\\
The proposed work is motivated by such an OL approach to MPC, which considers a generic algorithm Dynamic Mirror Descent (DMD) MPC, a framework that represents different MPC algorithms. DMD is reminiscent of the proximal update with a Bregman divergence that acts as a regularization to keep the current control distribution parameterized by $\eta_t$ at time $t$, close to the previous one. 
The second step of DMD uses the shift model $\Phi_t$ to anticipate the optimal decision for the next instant. \\
\\
The DMD-MPC proposes to use the shifted previous solution for shift model as  approximation to the current problem. The proposed methodology also aims to obtain an optimal policy for a finite horizon problem considering $H$-steps into the future using DMD MPC. \\
Denote the sequence of $H$ states and controls as $\mathbf{x}_t=(x_{t,0}, x_{t,1}, \dots, x_{t,H})$, and $\mathbf{u}_t=(u_{t,0}, u_{t,1}, \dots, u_{t,H-1})$, with $x_{t,0}=x_t$. The cost for $H$ steps is given by
\begin{align}
    C\left(\mathbf{x_t}, \mathbf{u_t}\right) = \sum^{H-1}_{h = 0} \gamma^h c(x_{t,h},u_{t,h}) + \gamma^H c_{H}(x_{t,H}) 
\end{align}
where, $c(x_{t,h}, u_{t,h}) = - r(x_{t,h}, u_{t,h})$ is the cost incurred (for the control problem) and $c_{H}(x_{t,H})$ is the terminal cost. Each of the $x_{t,h}, u_{t,h}$ are related by
\begin{equation}
\label{eq:innerloopdynamics}
    x_{t,h+1}  \sim f_\phi(x_{t,h},u_{t,h}), \quad h=0,1,\dots, H-1,
\end{equation}
with $f_\phi$ being the estimate of $f$. We will use the short notation $\mathbf{x_t}\sim f_\phi$ to represent \eqref{eq:innerloopdynamics}. It will be shown later that in a two-loop scheme, the terminal cost can be the value function obtained from the outer loop. 
\\
Now, by following the principle of DMD-MPC, for a rollout time of $H$, we sample the tuple $\mathbf{u_t}$ from a control distribution ($\pi_\eta$) parameterized by $\eta$. To be more precise, $\eta_t$ is also a sequence of parameters: 
\begin{equation*}
    \eta_t= (\eta_{t,0},\eta_{t,1}, \dots, \eta_{t,H-1})
\end{equation*}
 which yield the control tuple $\mathbf{u_t}$. Therefore, given the control distribution paramater $ \eta_{t-1}$ at round $t-1$,
we obtain $\eta_t$ at round $t$ from the following update rule:
\begin{align}
\label{eq:3}
    & \tilde{\eta}_{t} := \Phi_t (\eta_{t-1}) \nonumber \\
    & J(x_t,\tilde \eta_t) := \mathbb{E}_{\mathbf{u_t}\sim \pi_{\tilde \eta_t},  \mathbf{x_t}\sim f_\phi} \left [ C(\mathbf{x_t},\mathbf{u_t}) \right] \nonumber \\
    & \eta_t = \arg \min_{\eta} \ \left [ \alpha  \langle \nabla_{\tilde \eta_t} J(x_t,\tilde \eta_t) , \eta  \rangle + D_{\psi} (\eta \| \tilde \eta_t)  \right ] , 
\end{align}
where 
$J$ is the MPC objective/cost expressed in terms of $x_t$ and $\pi_{\tilde \eta_t}$, 
$\Phi_t$ is the shift model, $\alpha >0$ is the step size for the DMD, and $D_{\psi}$ is the Bregman divergence for a strictly convex function $\psi$. 
\\
Note that the shift parameter $\Phi_t$ is critical for convergence of this iterative procedure. Typically, this is ensured by making it dependent on the state $x_t$. In particular, for the proposed two-loop scheme, we make $\Phi_t$ dependent on the outer loop policy $\pi_{\theta_t}(x_t)$.
Also note that resulting parameter $\eta_t$ is still state-dependent, as the MPC objective $J$ is dependent on ${x_t}$.


With the two policies, $\pi_{\theta_t}$ and $\pi_{\eta_t}$ at time $t$, we aim to develop a synergy in order to leverage the learning capabilities of both of them. In particular, the ultimate goal is to learn them in ``parallel", i.e., in the form of two loops. The outer loop optimizes $\pi_{
\theta_t}$ and the inner loop optimizes $\pi_{\eta_t}$ for the MPC Objective. We discuss this in more detail in Section \ref{sec:methodology}.

\chapter{Methodology: Novel Framework \& Algorithms}
\label{sec:methodology}
In this chapter, we discuss a generic approach for combining model-free (Mf) and model-based (Mb) reinforcement learning (RL) algorithms through DMD-MPC. 
We define two new algorithms with  DMD MPC and RL: DeMo RL and DeMo Layer, one to improve sample efficiency of Off policy RL techniques while training and the other technique guides RL algorithms online for better policies. 

\section{Generalised Framework: DMD MPC \& RL}
\begin{figure}[t]
    \centering
    \includegraphics[width=\linewidth]{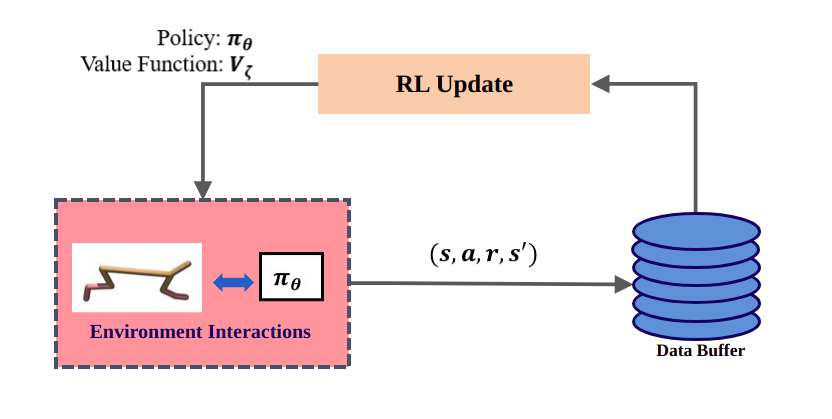}
    \caption{The Model Free Reinforcement Learning}
    \label{fig:methodmf}
\end{figure}

\begin{figure}[t]
    \centering
    \includegraphics[width=\linewidth]{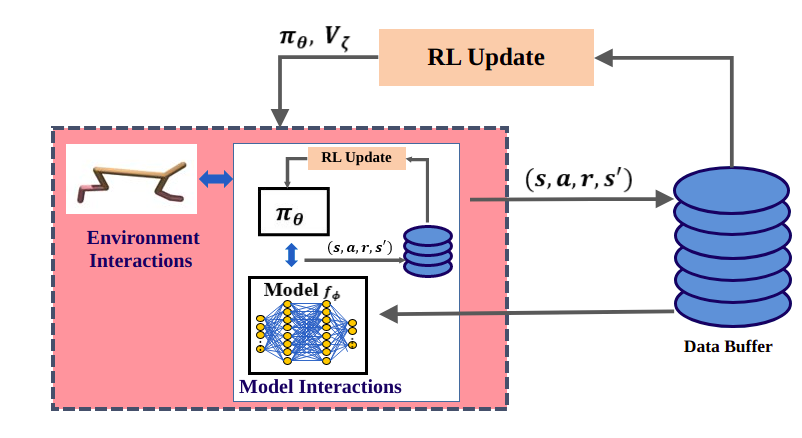}
    \caption{Model Based Reinforcement Learning.}
    \label{fig:methodmb}
\end{figure}

\begin{figure}[t]
    \centering
    \includegraphics[width=\linewidth]{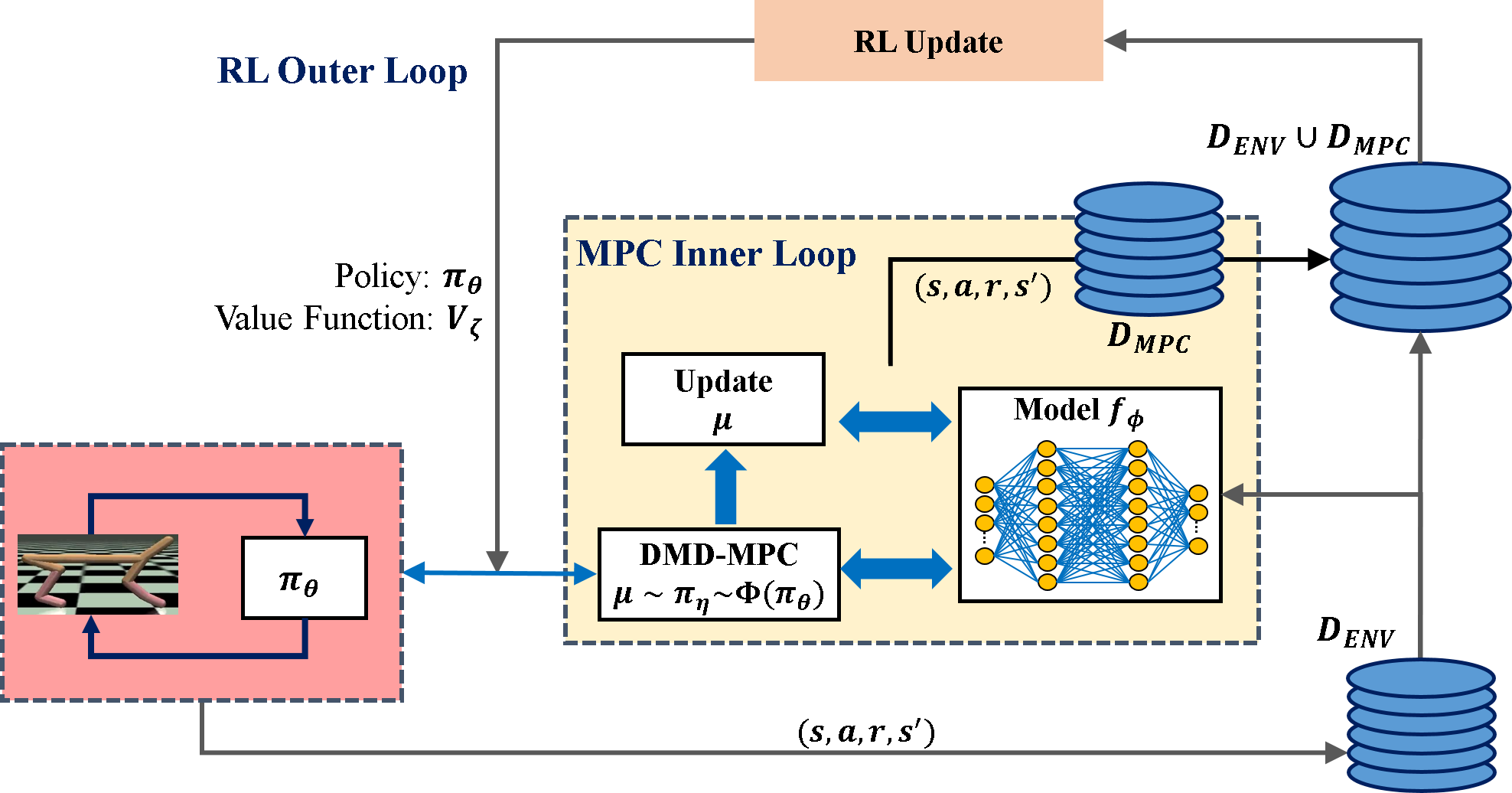}
    \caption{The proposed hierarchical structure of Dynamic-Mirror Descent Model-Predictive Reinforcement Learning (DeMoRL) with an inner loop DMD-MPC update and an outer loop RL update.}
    \label{fig:method}
\end{figure}

In classical Mf-RL, data from the interactions with the original environment are used to obtain the optimal policy parameterized by $\theta$. 
While the interactions of the policy are stored in memory buffer, $\mathcal{D}_{ENV}$, for offline batch updates, they are used to optimize the parameters $\phi$ for the approximated dynamics of the model, ${f}_\phi$. Such an optimized policy can then be used in the DMD-MPC strategy to update the control distribution, $\pi_{\eta}$. The controls sampled from this distribution are rolled out with the model, ${f}_\phi$, to collect new transitions and store these in a separate buffer $\mathcal{D}_{MPC}$. Finally, we 
%
update $\theta$ 
using both the data i.e., from the buffer $\mathcal{D}_{ENV} \cup \mathcal{D}_{MPC}$ 
via one of the off-policy approaches (e.g. DDPG \cite{lillicrap2015continuous}, SAC \cite{haarnoja2018soft}). 
In this work, we will demonstrate this using Soft Actor-Critic (SAC) \cite{haarnoja2018soft}. 
This gives a generalised hierarchical framework with two loops: Dynamic Mirror Descent (DMD) based Model Predictive Control (MPC) forming an inner loop and model-free RL in the outer loop. A graphical representation of Model Free RL, Model Based RL and the described framework are given in Figure~\ref{fig:methodmf}, Figure~\ref{fig:methodmb} and Figure~\ref{fig:method}.

There are two salient features in the two-loop approach:
\begin{itemize}
    \item[1.] At round $t$, we obtain the shifting operator $\Phi_t$ by using the outer loop parameter $\theta_t$. This is in stark contrast to the classical DMD-MPC method shown in \cite{wagener2019online}, wherein the shifting operator is only dependent on the control parameter of the previous round $\eta_{t-1}$.
    
\item[2.] Inspired by \cite{lowrey2018plan,morgan2021model}, the terminal cost $c_{H}(x_{t,H})=-~V_\zeta(x_{t,H})$ is the value of the terminal state for the finite horizon problem as estimated by the value function ($V_\zeta$, parameterized by $\zeta$) associated with the outer loop policy, $\pi_{\theta_t}$. This will efficiently utilise the model learned via the RL interactions and will in turn optimize $\pi_{\theta_t}$ with the updated setup. 
\end{itemize}


Since there is limited literature on theoretical guarantees of DRL algorithms, it is difficult to show convergences and regret bounds for the proposed two-loop approach. 
However, there are guarantees on regret bounds for dynamic mirror descent algorithms in the context of online learning \cite{hall2013dynamical}.
We restate them here using our notations for ease of understanding.
We reuse their following definitions:
\begin{align*}
G_J \triangleq \max_{\eta_t \in \mathcal{P}} \|\nabla J(\eta_t)\|, \quad M \triangleq \frac{1}{2} \max_{\eta_t \in \mathcal{P}} \|\nabla \psi(\eta_t)\|
\end{align*}
\begin{align*}
D_{max} \triangleq \max_{\eta_t,\eta_t' \in \mathcal{P}} D(\eta_t\|\eta_t'), \text{ and } \Delta_{\Phi_t} \triangleq \max_{\eta_t,\eta_t' \in \mathcal{P}} D(\tilde\eta_t\|\tilde\eta_t') - D(\eta_t\|\eta_t').   
\end{align*}
By a slight abuse of notations, we have omitted $x_t$ in the arguments for $J$. We have the following:
\begin{lemma}
\label{lem1}
Let the sequence $\tilde{\eta}_{t}$ be as in \eqref{eq:3}, and let ${\eta}_{t}$ be any feasible arbitrary sequence; then for the class of convex MPC objectives $J$, we have
\begin{equation*}
\begin{split}
    J\left(\tilde{\eta}_{t}\right)-J\left(\eta_{t}\right) \leq & \frac{1}{\alpha_{t}}\left[D\left(\eta_{t} \| \tilde{\eta}_{t}\right)-D\left(\eta_{t+1}\| \tilde{\eta}_{t+1}\right)\right]
    +\frac{\Delta_{\Phi_{t}}}{\alpha_{t}} + \frac{4\ M}{\alpha_{t}}\left\|\eta_{t+1}-\Phi_{t}\left(\eta_{t}\right)\right\|+\frac{\alpha_{t}}{2 \sigma} G_{J}^{2}
\end{split}
\end{equation*}
\end{lemma}


\begin{theorem}
\label{theo1}
    Given the shift operator $\Phi_{t}$ that is dependent on the outer-loop policy parameterised by $\theta$ at state $x_t$, the Dynamic Mirror Descent (DMD) algorithm using a diminishing step sequences $\alpha_{t}$ gives the overall regret with the comparator sequence $\eta_t$ as,
\begin{equation}
\mathcal{R}\left({\eta}_{T}\right) = \sum_{t=0}^{T} J\left(\tilde{\eta}_{t}\right)-J\left(\eta_{t}\right) \leq \frac{D_{\max }}{\alpha_{T+1}}+\frac{4 M}{\alpha_{T}} W_{\Phi_{t}}\left({\eta}_{T}\right)+\frac{G_{J}^{2}}{2 \sigma} \sum_{t=0}^{T} \alpha_{t}
\end{equation}
with
\begin{equation*}
W_{\Phi_{t}}\left({\eta}_{T}\right) \triangleq \sum_{t=0}^{T}\left\|\eta_{t+1}-\Phi_t(\eta_{t})\right\|.
\end{equation*}
Based on such a formulation, the regret bound is
$\mathcal{R}\left({\eta}_{T}\right)=O\left(\sqrt{T}\left[1+W_{\Phi_{t}}\left({\eta}_{T}\right)\right]\right)$.
\end{theorem}
   Proofs of both Lemma \ref{lem1} and Theorem \ref{theo1} are given in \cite{hall2013dynamical}. Theorem \ref{theo1} shows that the regret is bounded by $\|\eta_{t+1}-\Phi_t(\eta_{t})\|$, where the shifting operator $\Phi_t$ is dependent on the outer-loop policy.
However, this result is not guaranteed for non-convex objectives, which will be a subject of future work. 

Having described the main methodology, we will now study a widely used family of control distributions that can be used in the inner loop, the exponential family.

\subsection*{Exponential family of control distributions}

We consider a parametric set of probability distributions for our 
control distributions in the exponential family, given by natural parameters $\eta$, sufficient statistics $\delta$ and expectation parameters $\mu$ \cite{wagener2019online}.
Further, we set Bregman divergence in \eqref{eq:3} to the KL divergence, i.e.,
\begin{equation*} 
D_{\psi}\left( \eta \| \eta_{t}\right) \triangleq \operatorname{KL}\left(\pi_{\eta_t} \| \pi_{\eta}\right)
\end{equation*} 
After employing KL divergence, our $\eta_t$ update rule becomes:
\begin{equation}
\label{eq:7}
     \eta_t = \arg \min_{\eta \in \mathcal{P}} \ \left [ \alpha  \langle \nabla J(\mathbf{x_t},\tilde \eta_t) , \eta  \rangle +  \operatorname{KL}\left(\pi_{\tilde\eta_t} \| {\pi}_{\eta}\right)  \right ]
\end{equation}
The natural parameter of control distribution, $\tilde{\eta_t}$, is obtained with the proposed shift model $\Phi_t$ from the outer loop RL policy $\pi_{\theta_t}$
by setting the expectation parameter of $\tilde\eta_t$: $\mathbf{\tilde \mu_t}=  \pi_{\theta_t}(\mathbf{x_t})$. Note that we have overloaded the notation $\pi_{\theta_t}$ to map the sequence $\mathbf{x_t}$ to $\mathbf{\tilde \mu_t}$, which is the sequence of $\tilde \mu_{t,h} = \pi_{\theta_{t}}\left(x_{t,h}\right)$\footnote{Note that if the policy is stochastic, then $ \tilde{\mu}_{t,h} \sim \pi_{\theta_{t}}\left(x_{t,h}\right)$. This is similar to the control choices made in \cite[Algorithm 2, Line 4]{morgan2021model}.}.
Then, we have the following gradient of the cost:
\begin{align}
    \nabla_{\tilde \eta_t} J(\mathbf{x_t}, \tilde \eta_t) = \mathbb{E}_{\mathbf{u_t} \sim \pi_{\tilde \eta_t},\mathbf{x_t} \sim f_\phi} \left [ C(\mathbf{x}_t, \mathbf{u}_t)(\delta(\mathbf{u}_t) - \mathbf{\tilde{\mu}_t} )  \right ],
\end{align}
 where $\delta$ is the sufficient statistic, and 
 for our experiments we choose Gaussian distribution for control and $\delta(\mathbf{u_t}) := \mathbf{u_t}$. We finally have the following update rule for the expectation parameter \cite{wagener2019online}:
\begin{equation}
\label{eq:dmd-mpcexponentialfamily}
    \mathbf{\mu_t}=\left(1-\alpha\right) \mathbf{\tilde\mu_t}+\alpha \mathbb{E}_{\pi_{\tilde \eta_t}, f_\phi}\left[C\left(\mathbf{x_{t}}, \mathbf{u_t}\right) \mathbf{u_t}\right].
\end{equation}

Based on the data collected in the outer loop, the inner loop is executed via DMD-MPC as follows:
\begin{itemize}
\item[Step 1.] The shifting parameter $\Phi_t$ is obtained by using the outer loop parameter $\theta_t$. Now, considering $H$-step horizon, for $h=0,1,2, \dots, H-1$, obtain 
\begin{align}
\label{eq:innerDMDMPC1}
& \tilde \eta_{t,h}= \Sigma^{-1} \tilde \mu_{t,h}, \quad \tilde \mu_{t,h}=\pi_{\theta_t}(x_{t,h}) \\
\label{eq:innerDMDMPC2}
& u_{t,h} \sim \pi_{\tilde{\eta}_{t,h}}  \\
\label{eq:innerDMDMPC3}
& x_{t,h+1} \sim f_\phi(x_{t,h}, u_{t,h}). 
\end{align}
where 
$\Sigma$ represents the 
covariance for control distribution.
    \item[Step 2.] Collect 
        $\tilde \eta_t = ( \tilde \eta_{t,0}, \tilde \eta_{t,1}, \dots, \tilde \eta_{t,H-1}
    )$,
and apply DMD-MPC \eqref{eq:7} to obtain $\eta_t$.
\end{itemize}




\subsection*{MPC objective formulations}
Similar to the exponential family, we can use different types of MPC objectives. Specifically, we will be using the method of elite fractions that allows us to select only the best transitions. This is given by the following:
\begin{align}
J(\mathbf{x_t},\tilde \eta_t) := -\log \mathbb{E}_{\pi_{\tilde \eta_t}, f_\phi} \left[\boldsymbol{1}\left\{ C\left(\mathbf{x_t}, \mathbf{u_t}\right) \leq C_{t, \max }\right\}\right]
\end{align}
where we choose $ C_{t,max}$ as the top elite fraction from the estimates of rollouts. 
Alternative formulations are also possible, and, specifically, 
the objective used by the MPPI method in \cite{Williams2017imppi} is obtained by setting the following objective and $\alpha$ = 1 in \eqref{eq:dmd-mpcexponentialfamily} and for some $\lambda > 0$:
\begin{align}
\label{eq:mppiobjective}
J(\mathbf{x_t},\tilde \eta_t) =-\log \mathbb{E}_{\pi_{\tilde \eta_t}, f_\phi}\left[\exp \left(-\frac{1}{\lambda} C\left(\mathbf{x_t}, \mathbf{u_t}\right)\right)\right].
\end{align}
This shows that our formulation is more generic and some of the existing approaches could be derived with suitable choice: \cite{morgan2021model,Charlesworth2020SolvingCD} and \cite{NEURIPS2019_MBPO}. Table~\ref{table} shows the specific DMD-MPC algorithm and the corresponding shift operator used for each case.

\begin{table}[h]
\caption{ Mb-Mf algorithms as special cases of our generalised framework}
\label{table}
\centering
\begin{tabular}{c | c | c | c }
\hline
\hline
Mb-Mf Algorithm & RL & DMD-MPC & Shift Operator \\
\hline 
\hline
MoPAC & SAC  & MPPI & \textbf{Obtained from Mf-RL Policy} \\
\hline
TOPDM & TD3  & MPPI with CEM & \textbf{Left shift (obtained from the previous iterate)} \\
\hline
DeMoRL & SAC  & CEM & \textbf{Obtained from Mf-RL Policy} \\
\hline
\hline
\end{tabular}
\end{table}



\section{DeMo RL Algorithm}

DeMoRL algorithm derives from other Mb-Mf methods in terms of learning dynamics and follows a similar ensemble dynamics model approach. We have shown it in Algorithm \ref{algorithm}. There are three parts in this algorithm: Model learning, Soft Actor-Critic and DMD-MPC. We describe them below.

\textbf{Model learning.} The functions to approximate the dynamics of the system are $K$-probabilistic deep neural networks \cite{NEURIPS2019_deepdynamicmodel} 
 cumulatively represented as 
$\{f_{\phi_1}, f_{\phi_2}, \dots, f_{\phi_K}\}$. Such a configuration is believed to account for the epistemic uncertainty of complex dynamics and overcomes the problem of over-fitting generally encountered by using single models \cite{NEURIPS2018_handfultrials}. 

\textbf{SAC.} Our implementation of the proposed algorithm uses Soft Actor-Critic (SAC) \cite{haarnoja2018soft} as the model-free RL counterpart. Based on principle of entropy maximization, the choice of SAC ensures sufficient exploration motivated by the soft-policy updates, resulting in a good approximation of the underlying dynamics. 

\textbf{DMD-MPC.} Here, we solve for $\mathbb{E}_{\pi_{\tilde \eta_t}, f_\phi}\left[C\left(\mathbf{x_{t}}, \mathbf{u_{t}}\right) \mathbf{u_{t}}\right]$ using a Monte-Carlo estimation approach. For a horizon length of $H$, we collect $M$ trajectories using the current policy $\pi_{\theta_t}$ and the more accurate dynamic models from the ensemble having lesser validation losses. For all trajectories, the complete cost is calculated using a deterministic reward estimate and the value function through (2). After getting the complete state-action-reward $H$-step trajectories we execute the following based on the CEM \cite{pourchot2018cemrl} strategy:
\begin{itemize}
    \item[Step 1] Choose the $p\%$ elite trajectories according to the total $H$-step cost incurred. We set $p=10\%$ for our experiments, and denote the chosen respective action trajectories and costs as $U_{elites}$ and $C_{elites}$ respectively. Note that we have also tested for other values of $p$, and the ablations are shown in the Appendix attached as supplementary.
    
    \item[Step 2] Using $U_{elites}$ and $C_{elites}$ we calculate $\mathbb{E}_{\pi_{\tilde \eta_t}, f_\phi}\left[C\left(\mathbf{x_{t}}, \mathbf{u_{t}}\right) \mathbf{u_{t}}\right]$ as the reward weighted mean of the actions i.e.
    \begin{align}
    \label{eq:elitecalc1}
        \mathbf{g_t} = \frac{\sum_{i \in elites} {C}_{i} \ U_{i}}{\sum_{i \in elites} {C}_{i}}
    \end{align}
    
    \item[Step 3] Finally, we update the current policy actions, $\mathbf{\tilde\mu_t} = \pi_{\theta_t}(\mathbf{x_t})$ according to \eqref{eq:dmd-mpcexponentialfamily} as
    \begin{align}
    \label{eq:elitecalc2}
        \mathbf{\mu_{t}}=\left(1-\alpha\right) \mathbf{\tilde\mu_t} +\alpha \mathbf{g_t}.
    \end{align}
\end{itemize}

\begin{algorithm}[H]
    \caption{DeMoRL Algorithm}
    \label{algorithm}
    \begin{algorithmic}[1]

    \STATE $ \text {Initialize SAC and Model: } \phi, \text{SAC and Environment Parameters} $
    \STATE $ \text {Initialize memory buffers: } D_{ENV}, D_{MPC}$ 
    
    \FOR{each iteration}
    \STATE $D_{ENV} \leftarrow D_{ENV} \cup\left\{x, u, r, x'\right\}, u \sim \pi_{\theta}\left(x\right)$

    \FOR {each model learning epoch}
    \STATE $\text { Train model } f_{\phi} \text{ on } D_{ENV} \text{ with loss}: J_{\phi} = \|x' - f_\phi(x,u)\|_2$
    \ENDFOR
    
    \FOR{ each DMD-MPC iteration}
    \STATE $\text {Sample } x_{t,0} \text { uniformly from } D_{ENV} $
    \STATE $\text{Simulate M trajectories of H steps horizon: }$ \eqref{eq:innerDMDMPC1}, \eqref{eq:innerDMDMPC2} and \eqref{eq:innerDMDMPC3}
    \STATE $\text{Perform CEM to get optimal action sequence: } \mathbf{\mu_t}$ \eqref{eq:elitecalc1} and \eqref{eq:elitecalc2}
    \STATE $\text{Collect complete trajectory: } \mathbf{x_t} \sim f_\phi(x_{t,0}, \mathbf{\mu_t})$ 
    \STATE $ \text {Add all transitions to } D_{MPC}: D_{MPC} \leftarrow D_{MPC} \cup\left\{x_{t,h}, u_{t,h}, \hat{r}_{t,h}, x_{t,h+1}\right\}$
    \ENDFOR
    \FOR {each gradient update step} 
    \STATE Update SAC parameters using data from $D_{ENV} \cup D_{MPC}$ 
    \ENDFOR
    \ENDFOR
    \end{algorithmic}
\end{algorithm}

\section{DeMo Layer}

We consider the special case of above generalised framework where the outer loop RL is not updated and is already trained till convergence. At a state, the RL policy gives distribution over actions. With the shift model obtained from the trained RL the equation \ref{eq:7} now has fixed shift model.
\begin{equation}
\label{eq:7}
     \eta_t = \arg \min_{\eta } \ \left [ \alpha  \langle \nabla J(x_{t},\tilde \eta_t) , \eta  \rangle +  \operatorname{KL}\left(\pi_{\tilde\eta_t} \| {\pi}_{\eta}\right)  \right ]
\end{equation}
The $\tilde{\eta_t}$,is obtained from RL policy  $\pi_{\theta}$ by setting
\begin{equation*}
   \mu_t=  \pi_{\theta}(\mathbf{x_t}) 
\end{equation*}
The updated policy $\pi_\eta $ is optimal both in terms of long term expected reward and short term horizon based cost.Following derivations from previous section we have the closed form expression for action with Gaussian distribution for control policy,
        \[u_{t} =\left(1-\gamma_{t}\right) u_{t,RL}+\gamma_{t} \mathbb{E}_{\pi_{\tilde{\theta}_{t}}, f}\left[C_{t}\left(\hat{x}_{t}, \hat{\boldsymbol{u}}_{t}\right) \hat{u}_{t,}\right]\]
        \\
This gives an action which is a convex combination of RL action and the action that are good according to the current cost $J$
\begin{figure}[t]
    \centering
    \includegraphics[width=\linewidth]{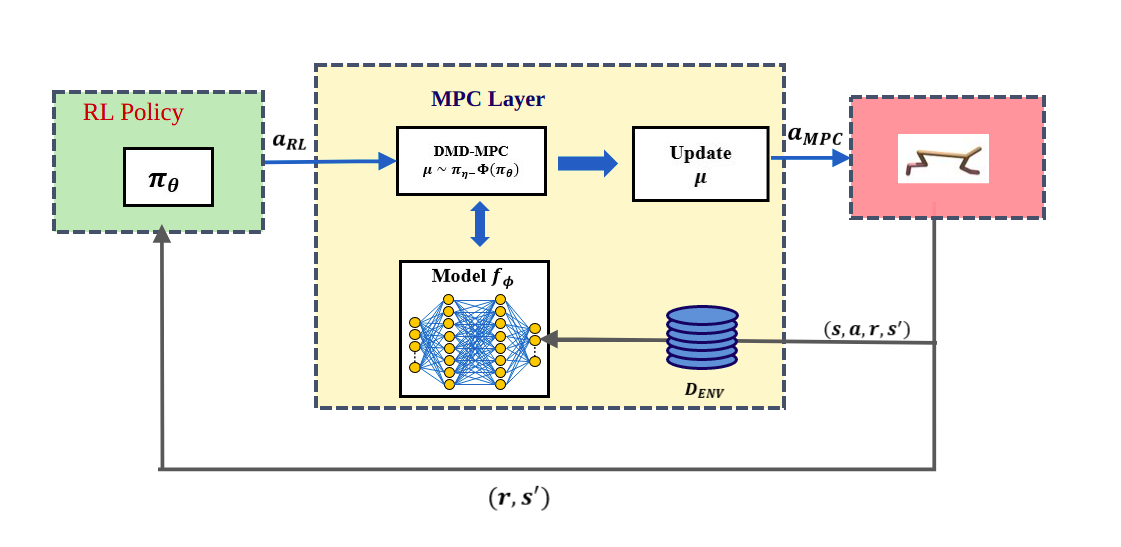}
    \caption{The proposed DeMo Layer with an inner loop DMD-MPC update to guide outer loop RL.}
    \label{fig:demomethod}
\end{figure}
We sample an action from the updated policy $\pi_\eta$ and apply it to the real environment unlike previous case, thus guiding the RL Policy End to End.A graphical representation of the DemoLayer framework is given in Figure \ref{fig:demomethod}.
We describe the three parts of DeMo Layer, here:  Model learning, ARS and DMD-MPC.

\textbf{Model Learning}:

\hspace{0.3cm} \textbf{Stoch}:Model Dynamics is learnt using Feed forward Neural Networks \cite{nagabandi2018neural} 

\hspace{0.3cm} \textbf{Cartpole}:We used the Model given in Open AI gym with biased length for the MPC

\textbf{ARS}: It is a linear deterministic policy and we have implemented this with modification as in Algorithm \ref{arsalgorithm} for faster convergence.

\textbf{DMD-MPC.}: We use the similar strategy as described in DeMo RL.

\chapter{Experimental Results}
\label{sec:result}

In this section, we implement the two-loop hierarchical framework as explained in the previous sections for the DeMo RL and DeMo Layer Setting. We will compare the DeMo RL with the existing approaches MoPAC \cite{morgan2021model} and MBPO \cite{NEURIPS2019_MBPO} on the benchmark MuJoCo control environments. We test the DeMo Layer for cartpole and custom build Quadruped Stoch.

\section{DeMo RL: Results and Comparison}

Several experiments were conducted on the MuJoCo \cite{todorov2012mujoco} continuous control tasks with the OpenAI-Gym benchmark and the performance was compared with recent related works MoPAC \cite{morgan2021model} and MBPO \cite{NEURIPS2019_MBPO}. First, we discuss the hyperparameters used for all our experiments and then the performance achieved in the conducted experiments

\begin{figure}[h]
\centering
\begin{subfigure}[width=8cm]{.4\textwidth}
  \centering
  \includegraphics[width=\linewidth]{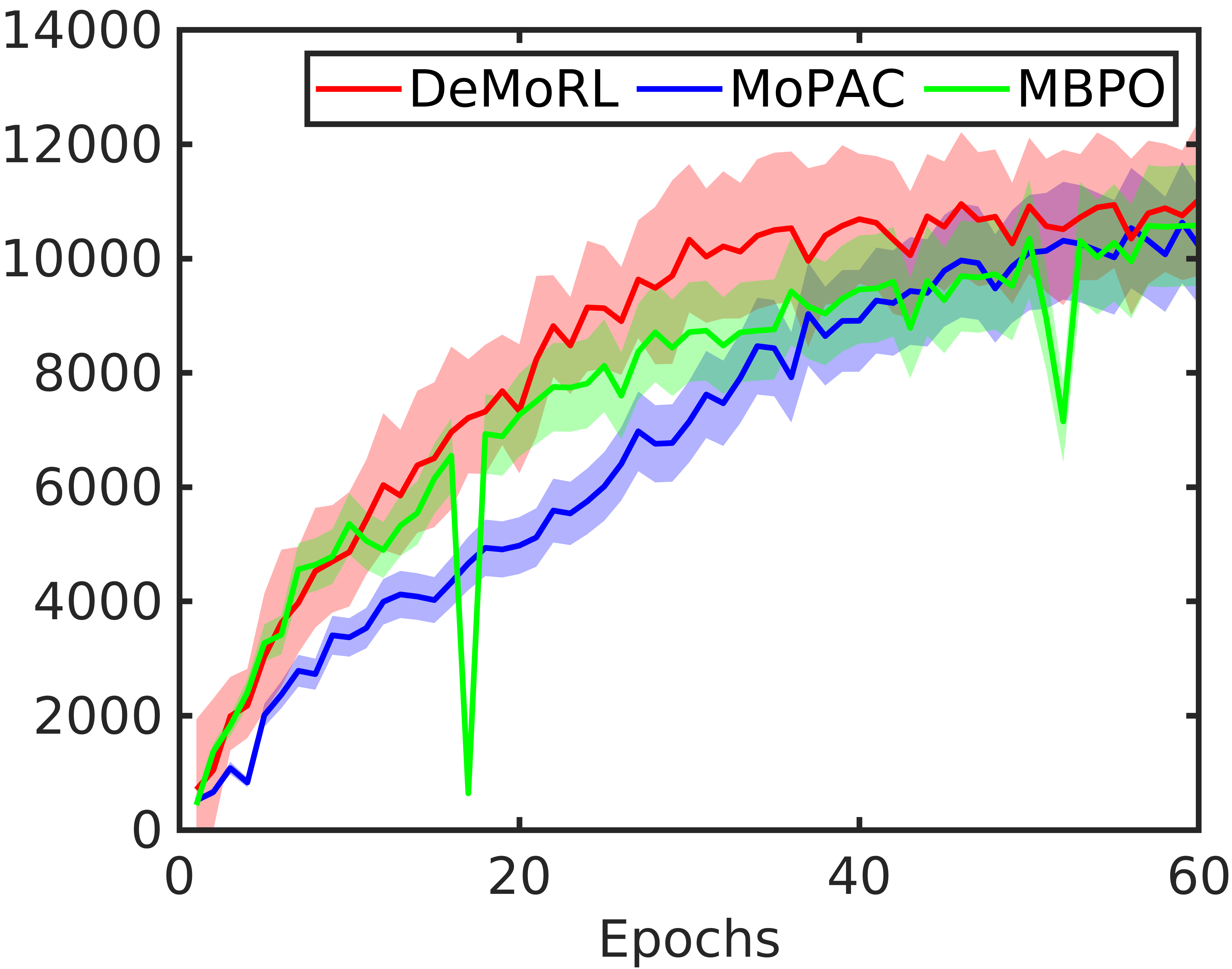}
  \caption{HalfCheetah-v2}
\label{fig:rewcheetah}
\end{subfigure}
\begin{subfigure}{.4\textwidth}
  \centering
  \includegraphics[width=\linewidth]{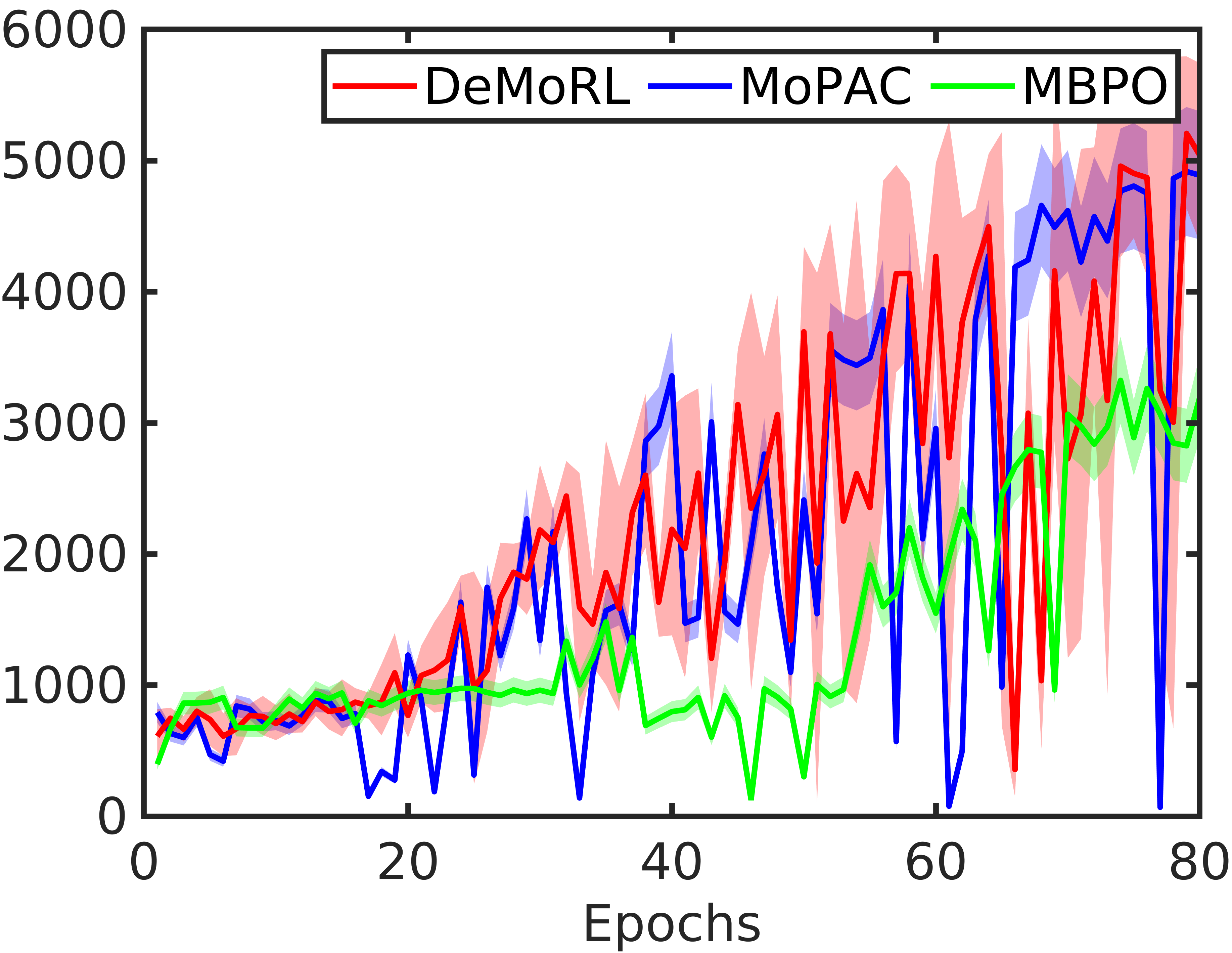}
  \caption{Ant-v2}
\label{fig:rewant}
\end{subfigure}
\begin{subfigure}{.4\textwidth}
  \centering
  \includegraphics[width=\linewidth]{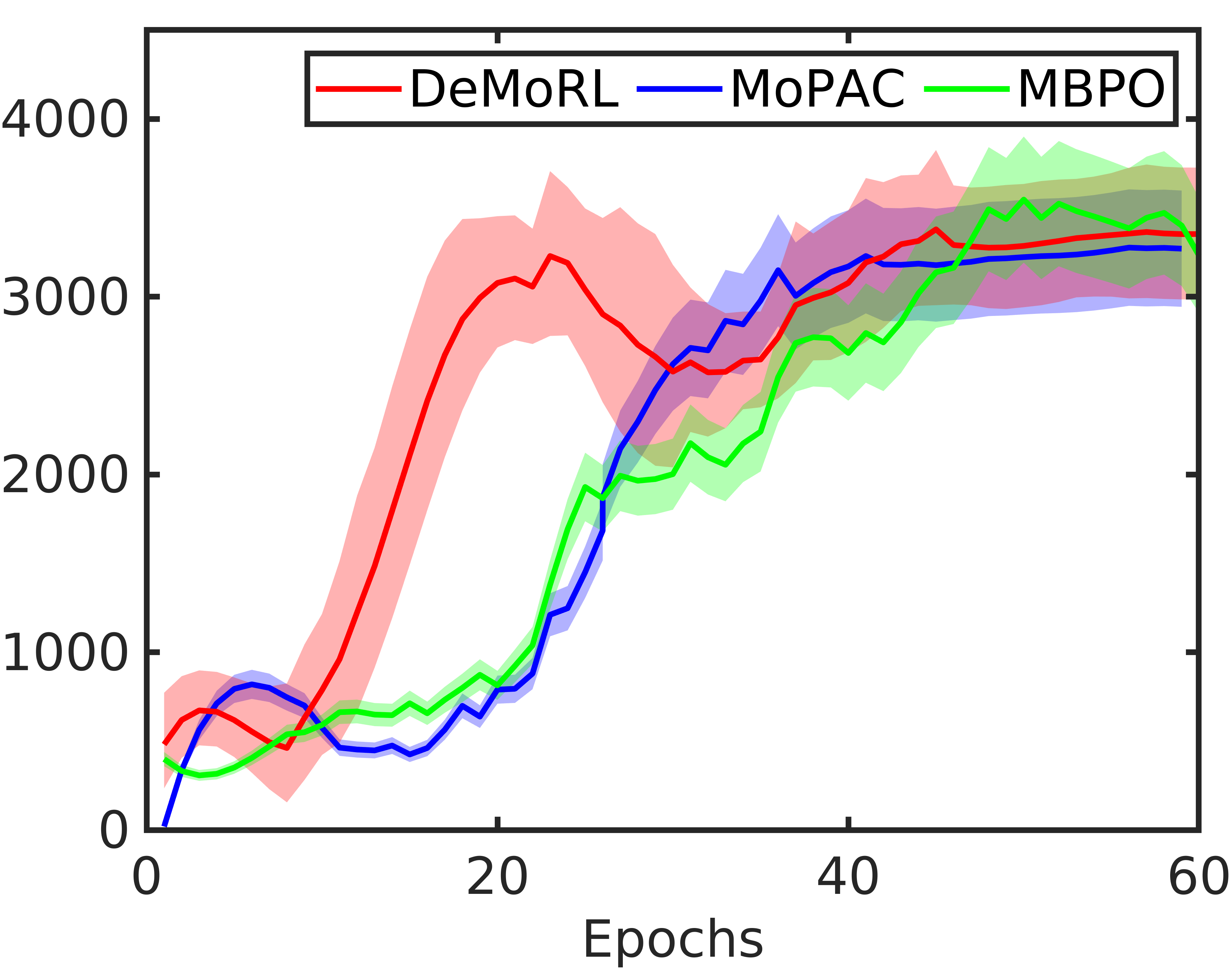}
  \caption{Hopper-v2}
\label{fig:rewhopper}
\end{subfigure}
\begin{subfigure}{.4\textwidth}
  \centering
  \includegraphics[width=\linewidth]{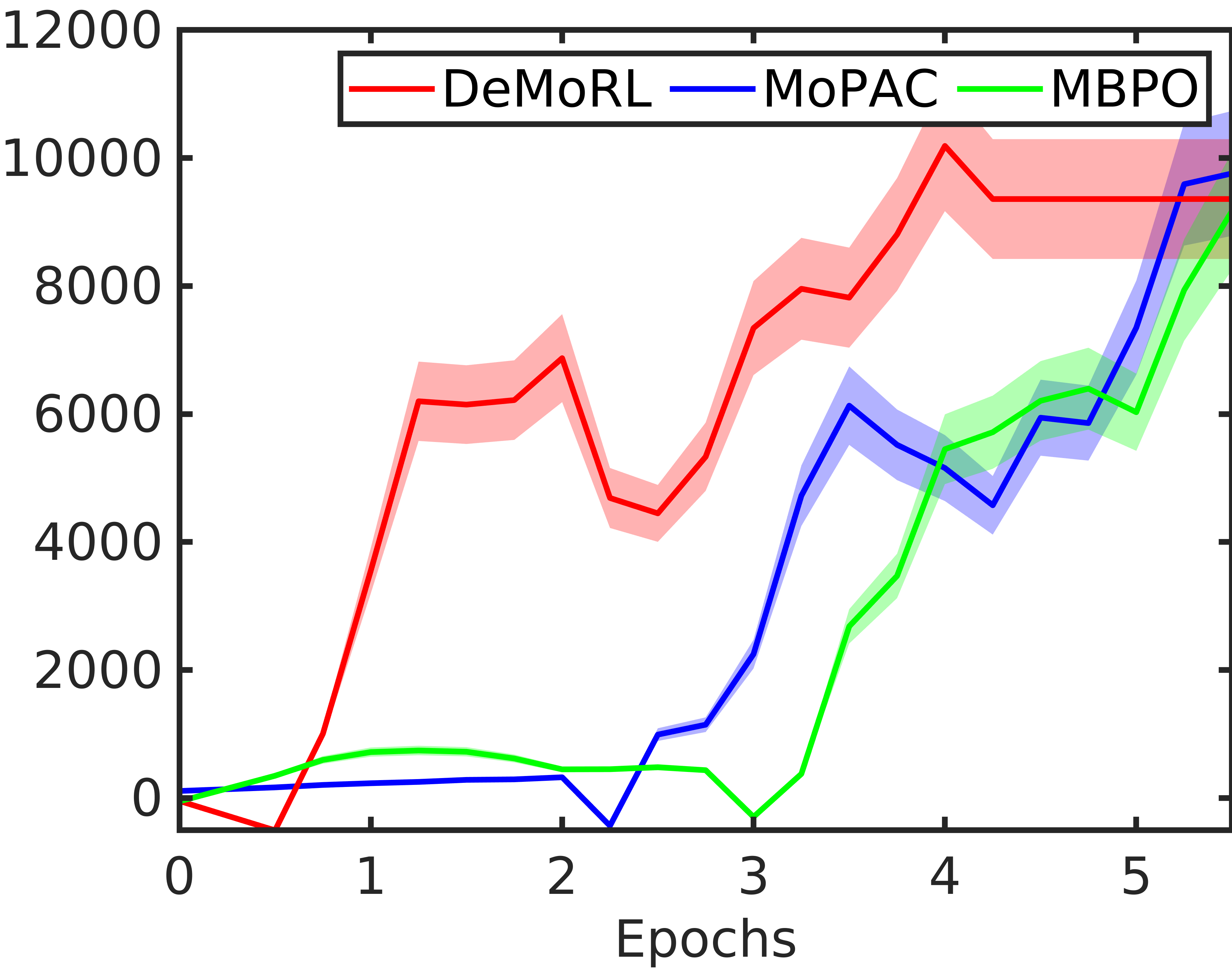}
  \caption{InvertedDoublePendulum-v2}
\label{fig:rewidp}
\end{subfigure}
\caption{Reward Performance of DeMoRL algorithm over other model based algorithms: MoPAC and MBPO.}
\label{fig:results}
\end{figure}

As the baseline of our framework is built upon MBPO implementation, we use the same hyperparameters for our experiments and both the algorithms. We compare the results of three different seeds and the reward performance plots are shown in Figure \ref{fig:results}. For the inner DMD-MPC loop we choose a constant horizon length of $15$ and perform $100$ trajectory rollouts. With our elite fraction as $10\%$, the updated model-based transitions are added to the MPC-buffer. This process is iterated with a batch-size of $10,000$ thus completing the DMD-MPC block in Algorithm \ref{algorithm}. Inspired from MoPAC and MBPO, the number of interactions with the true environment for SAC were kept constant to $1,000$ for each epoch.
\\

For HalfCheetah-v2, Hopper-v2 and InvertedDoublePendulum-v2, we clearly note an accelerated progress with approximately $30\%$ faster rate in the reward performance curve. Whereas in Ant-v2, our rewards were comparable with MoPAC but still significantly better than MBPO. 
Our final rewards are eventually the same as achieved by MoPAC and MBPO, however the progress rate is faster for all our experiments. 
and requires lesser number of true environment interactions. Furthermore, all the experiments were conducted with the same set of hyperparameters, thus tuning them individually might give better insights. Table \ref{table} shows the empirical analysis of the acceleration achieved by DeMoRL.

\begin{table}[h]
\caption{Mean Reward Performance Analysis of DeMo RL}
\label{table}
\centering
\begin{tabular}{|c | c | c | c | c|}
\hline
Environment & Epochs & 20 & 40 & 60 \\
\hline
\multirow{4}{*}{HalfCheetah-v2} & DeMoRL  & \textbf{7333} & 10691 & 11037\\
& MoPAC  & 4978 & 8912 & 10212\\
& MBPO  & 7265 & 9461 & 10578\\
\hline
\multirow{4}{*}{Ant-v2} & DeMoRL  & \textbf{984.0} & 2278.4 & 3845.5\\
& MoPAC  & 593.6 & 2337.3 & 3649.5\\
& MBPO  & 907.5 & 1275.6 & 1891.9\\
\hline
\multirow{4}{*}{Hopper-v2} & DeMoRL  & \textbf{3077.3} & 3077.5 & 3352.4\\
& MoPAC  & 789.9 & 3137.9 & 3270.2\\
& MBPO  & 813.9 & 2683.5 & 3229.9\\
\hline
\end{tabular}
\end{table}


Here, we not only show a generic formulation of the DMD-MPC, but also demonstrate how new types of objectives can be obtained and further improvements can be made. 
As shown in the table, we perform better than or at least as good as MoPAC, which uses 
information theoretic model predictive path integrals (i-MPPI) \cite{Williams2017imppi}, a special case of our setup as shown in \eqref{eq:mppiobjective}. 
The MPPI formulation uses all the rollouts to calculate the action sequence while the CEM uses elite rollouts, which contributed to the accelerated progress. 


Here we show a detailed study on the elite percentage
%
referring to the previous steps, after getting complete state-action-reward $H$-step trajectories,
we execute Steps 1 to 3 in page $16$.
    
    

\begin{figure}[h!]
    \centering
    \includegraphics[width=0.8\linewidth]{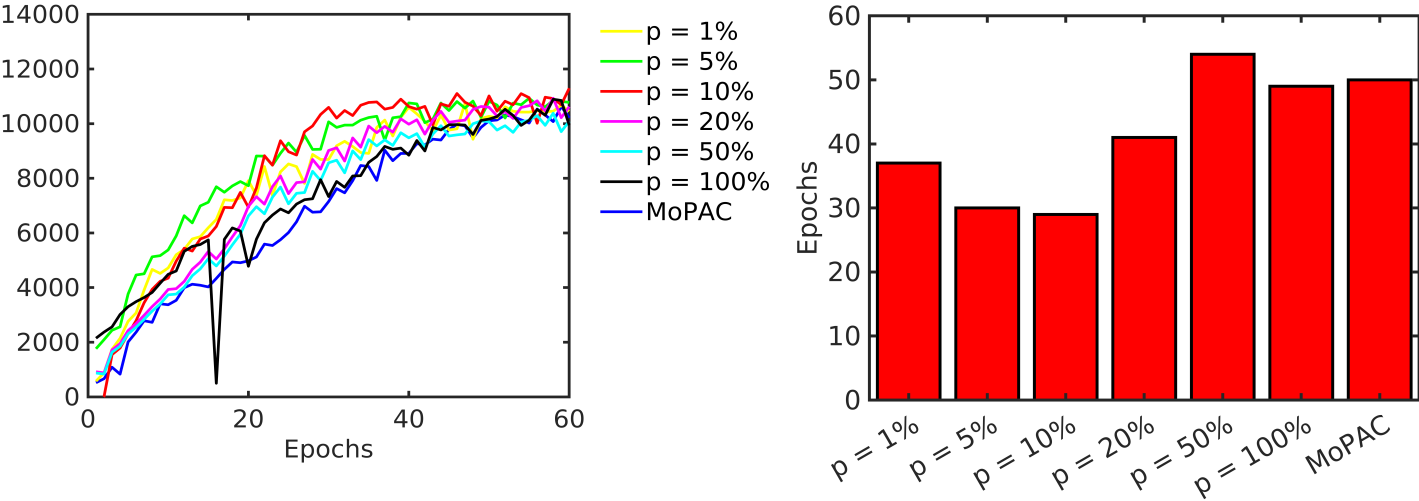}
    \caption{Ablation study for elite percentage: Reward performance curve (left) and Acceleration analysis as epochs to reach 10000 rewards (right)}
    \label{fig:ablation}
\end{figure}

Given the sequence of controls $\mathbf{\mu_t}$, we collect the resulting trajectory and add them to our buffer. Therefore, the quality of $\mathbf{\mu_t}$ is a significant factor affecting the quality of data used for the outer loop RL-policy. The selection strategy being CEM, a quality metric is dependent on the choice of elite fractions $p$. 
We perform an ablation study for $6$ values of $p = 1, 5, 10, 20, 50$ and $100\%$ on HalfCheetah-v2 OpenAI gym environment. The analysis was performed based on the reward performance curves as shown in Fig.~\ref{fig:ablation} (left). Additionally, we realize the number of the epochs required to reach a certain level of performance as a good metric to measure acceleration achieved. Such an analysis is provided in Fig.~\ref{fig:ablation} (right). We make the following observations:
\begin{itemize}
    \item Having a lesser value of $p$ might ensure that learned dynamics is exploited the most, but decreases the exploration performed in the approximated environment.
    \item Similarly, having higher value of $p$ on the other hand will do more exploration using a ``not-so-perfect" policy and dynamics. 
\end{itemize}
Thus, the elite fraction balances between exploration and exploitation. 

\section{DeMo Layer: Results for Cartpole and Stoch}
We have conducted experiments on two different environments. \\
\textbf{Cartpole}:
A linear policy is trained on Cartpole using ARS and there exists no linear policy that could acheive swing up and balance. Showed that DeMo Layer could guide the linear policy to acheive swing up and balance on cartpole.\\
\textbf{Stoch}:
Stoch2, a quadruped Robot is trained using the linear approach given in \cite{paigwar2020robust}.
With neural network approximation to the Model Dynamics, DeMo Layer is implemented on Stoch to learn robust walking for episode length of 500.\\

\begin{table}[h]
\caption{Reward Performance Analysis of DeMo Layer }
\label{table}
\centering
\begin{tabular}{|c | c | c |}
\hline
Environment & Linear Policy & Linear Policy with DeMo Layer \\
\hline
Cartpole & 1400  & \textbf{1700}\\
\hline
Stoch2 & 1500  & \textbf{1850}\\
\hline

\end{tabular}
\end{table}

\begin{table}[h]
\caption{Hyper parameters used for DeMo Layer }
\label{table}
\centering
\begin{tabular}{|c | c | c |}
\hline
Environment & Horizon & Sampled Trajectories \\
\hline
Cartpole & 120  & 90 \\
\hline
Stoch2 & 20  & 200 \\
\hline

\end{tabular}
\end{table}

The simulation results for cartpole and stoch could be found here \footnote{https://github.com/soumyarani/End-to-End-Guided-RL-using-Online-Learning}

\chapter{Conclusion and Future Work}
\label{sec:conclusion}
We proposed a generic framework using a novel combination of DMD MPC with Model Free RL. Different choices gives existing Mb-Mf Approaches with added flexibility to use off the shelf RL algorithms and ease to define new DeMo RL Algorithms with different MPC techniques. We have investigated the role of leveraging the model-based optimisation with online learning to accelerate model-free RL algorithms using DeMo RL and we have given an end to end algorithm DeMo Layer that could be used with RL algorithms for online adaption. While several methods use model predictive rollouts in Mb-Mf approaches, we give a generalization using this framework for different MPC Controllers with Mf-RL algorithms. 
\\

Further, mirror-descent ensures that the MPC policy and the RL policy are proximal enough so that transitions from MPC do not suffer from distributional shift. Our analysis shows that our formulation is generic and achieves asymptotic performance of Mf-RL algorithms as the underlying online learning tracks for the best policy and ultimately converges to the target Mf-RL policy. Further the empirical study shows, the policy is learnt substantially faster than prior Mf-Mb methods.
\\

Future work would involve testing on the Real Robot and regret analysis of proposed hierarchical approach. We would like to apply DeMo RL on other complex environments like dexterous hand manipulation. We would also like to extend theDeMo Layer to Safe RL Settings.


\bibliographystyle{plainnat}

\end{document}